\documentclass[10pt, conference, letterpaper]{IEEEtran}

\usepackage{amsmath,amsfonts,stackrel}
\usepackage{epsfig}
\usepackage{epstopdf}
\usepackage{multirow}
\usepackage{hhline}
\usepackage{verbatim}
\usepackage{lipsum}
\usepackage{graphicx,subfigure}
\usepackage{enumitem}
\usepackage{textcomp}
\usepackage{relsize}
\usepackage{multicol}
\usepackage{multirow}
\usepackage{xcolor}
\usepackage[c3]{optidef}
\usepackage{tabularx,ragged2e}
\usepackage[bottom=1in, right=0.56in, left=1.4cm, top=0.68in]{geometry}
\usepackage{array}
\usepackage{algorithmic}

\usepackage[ruled,vlined]{algorithm2e}

\usepackage{enumitem}
\usepackage{xcolor,soul}

\newcolumntype{C}{>{\centering\arraybackslash}X}
\newcolumntype{P}[1]{>{\centering\arraybackslash}p{#1}}

\newtheorem{property}{Property}

\definecolor{cadmiumgreen}{rgb}{0.12, 0.8, 0.17}
\usepackage{times}
\usepackage{fancyhdr}

\newcommand{\Hc}{{\mathcal{H}}}
\newcommand{\Nc}{{\mathcal{N}}}
\newcommand{\Ac}{{\mathcal{A}}}
\newcommand{\bsigma}{{\boldsymbol{\sigma}}}

\newcommand{\Eq}[1]{(\ref{eq:#1})}

\begin{document}
\title{Resource-Efficient Sensor Fusion\\ via System-Wide Dynamic Gated Neural Networks\vspace{-0pt}}

\author{C. Singhal$^{1}$, Y.~Wu$^{2}$, F.~Malandrino$^{3,4}$,  S.~Ladron de Guevara Contreras$^{2}$, M.~Levorato$^{2}$, C.~F.~Chiasserini$^{5,4,6}$\\
1: INRIA, France -- 2: 
UC Irvine, USA -- 3: CNR-IEIIT, Italy\\ 4: CNIT, Italy -- 5: Politecnico di Torino, Italy  -- 6: Chalmers University of Technology, Sweden
\vspace*{-0.5 cm}
}
\maketitle

\begin{abstract}
Mobile systems will have to support multiple AI-based applications, each leveraging heterogeneous  data sources through DNN architectures collaboratively executed within the network. To minimize the cost of the AI inference task subject to requirements on latency,  quality, and -- crucially --  {\em reliability} of the inference process, it is vital to optimize (i) the set of sensors/data sources and (ii) the DNN architecture, (iii) the network nodes executing sections of the DNN, and (iv) the resources to use. To this end, we leverage dynamic gated neural networks with branches, and propose a novel algorithmic strategy called  Quantile-constrained Inference (QIC), based upon quantile-Constrained policy optimization. QIC  makes joint, high-quality, swift decisions on all the above aspects of the system, with the aim to minimize inference energy cost. We remark that this is the first contribution connecting gated dynamic DNNs with infrastructure-level decision making. We evaluate QIC using a dynamic gated DNN with stems and branches for optimal sensor fusion and inference, trained on the RADIATE dataset offering Radar, LiDAR, and Camera data, and real-world wireless measurements. Our results confirm  that QIC matches the optimum and outperforms its alternatives
by over 80\%.  
\end{abstract}


\section{Introduction}
\label{sec:intro}

Several emerging mobile applications are powered by artificial intelligence (AI) algorithms, often processing and fusing the data produced by multiple sensors and {\em data sources}. Many of such algorithms take the form of deep neural networks (DNNs). An intriguing class of neural architectures include {\em branches}, also known as early exits~\cite{survey}. These architectures are dynamic, as the execution adapts to the characteristics of the input: the branches are sequentially executed until a satisfactory output is produced. 
The state of the art in neural networks evolved past early exits, and recently models with more structured and complex adaptability were developed. Specifically, these models are articulated into sections, connected by neural structures called {\em gates}. The gates directly control the internal routing of information based on optimal execution strategies learned during training~\cite{han2021dynamic}. In their simplest form, gates pre-select a full model, or a set of models in mixture of experts settings, based on simple features of the input. In more complex instances, such as that in~\cite{malawade2022hydrafusion} and the one considered in this paper, gated models are crafted as the composition of stems extracting features from a diverse set of sensors, and branches that process these intermediate features to produce a final output. The gates, then, control the activation of the stems, and the way features referring to different input sensors are fused and analyzed, and, in the model developed in this paper, the modality of sensor fusion. This class of models, whose design and training are highly non-trivial, perform a dynamic and context-aware form of sensor fusion.

Current literature develops and analyzes these models in isolation, and considering execution on a single device.
In contrast, {\bf this paper represents the first contribution considering dynamic gated models in the context of layered computing/communication infrastructures -- i.e., those composed of interconnected mobile nodes and edge servers.} In such setting, these architectures represent a significant opportunity, as they enable a flexible and efficient allocation of computing and communication load across different layers of the system. Importantly, as the gates control the structure of the neural network model and the use of data sources (that is, the combination of {\em stems} and {\em branches} mapping the input to the output) in response to input characteristics, the whole resource allocation strategy becomes {\bf context and input aware}. We, thus, define the gates as connected to an infrastructure-level orchestrator that directly controls the activation of the DNN sections, and the resources on which they will be executed. That is, the model can be split at the gates and executed on different system resources (mobile nodes and edge server). Such characteristics and interplay between the inner neural network structure  and the operations of the infrastructure becomes especially important when considering multiple applications -- each with different accuracy and latency requirements -- coexisting on the same resources.

The use of such architectures therefore 
results in (potentially) effective data analysis, improved efficiency, and lower costs. At the same time, achieving these results requires swift, high-quality, and {\em joint} decisions about such diverse aspects as:
1) the data sources to leverage for each application;
    2) the DNN sections (i.e., stems and branches) to use;
    3) the network nodes where stems and branches shall run;
    4) the computation and communication resources to devote to each application and at each node.
All these decisions have to be made with the goal of minimizing the cost (e.g., energy) of inference, subject to inference quality (e.g., accuracy) and  latency requirements.
We remark that each data source is connected to a different {\em stem} of the DNN, which  produces features that are then processed by a {\em branch}  tailored to the quantity and type of the data the source produces (e.g., 2D or 3D images), to generate the final inference output. 
Also, importantly, we do not express inference quality requirements  in terms of average/expected values (e.g., the expected accuracy), but rather in terms of a target {\em quantile} thereof (e.g., the 90th percentile of accuracy). On the one hand, this reflects the time- and performance-critical nature of many modern applications, which need to make correct decisions with a {\em guaranteed} (high) probability. At the same time, the added flexibility of targeting arbitrary quantiles also accommodates the opposite scenario: applications for which occasional failures are acceptable and/or have limited consequences can trade some accuracy for a lower cost.

On the negative side, making all the decisions mentioned above while accounting for inference quality quantiles is dauntingly complex, for three main reasons. The most straightforward is the {\em scale} of the problem, e.g., the very large number of alternatives to consider. Furthermore, the combinatorial {\em structure} of the problem (coming from, among other things, the presence of a discrete set of stems and branches  to choose from) rules out direct optimization approaches. Finally, and most important, the {\em nature} of the problem itself is utterly new, which makes existing approaches, like those discussed in Sec.~\ref{sec:relwork}, impossible to apply to our scenario.

In this work, we fill this gap by proposing a new solution strategy called {\em Quantile-constrained Inference based on quantile-Constrained policy optimization} (QIC). QIC makes decisions that are: (i) {\em joint}, as they account for data sources, stems, branches, network nodes and resources to use for each application; (ii) {\em dependable}, as they can guarantee an arbitrary value of an arbitrary quantile of the inference quality; (iii) {\em efficient and effective}, as near-optimal choices are made in  polynomial time.
Our contributions in this work go beyond QIC itself and can be summarized as follows:

\noindent
    \textbullet\, We provide (Sec.~\ref{sec:description}) a comprehensive description of the applications and scenario we target, along with (Sec.~\ref{sec:model}) a synthetic, yet expressive,  model accounting for all the most relevant features of the system;

    \noindent
\textbullet\,  We develop (Sec.\,\ref{sub:DNN_architecture}) a new instance of dynamic gated neural model for object detection performing sensor fusion on various types of data. Notably, the model's internal configuration is connected to infrastructure-level decisions, and its sections can be deployed over different system's resources; 

\noindent
  \textbullet\,  We develop the QIC solution framework, which innovatively applies quantile-constrained optimization on the dynamic graph representing the system evolution over time, and characterize its  efficiency (Sec.~\ref{sec:solution});

 \noindent 
    \textbullet\,  We build a  real-world  reference scenario, including state-of-the-art DNNs and real-world wireless measurements (Sec.~\ref{sec:refscen});

  \noindent  
\textbullet\,  In this scenario, we study QIC's performance (Sec.~\ref{sec:peva}), finding it to significantly outperform state-of-the-art approaches (80\% and 50\% reduction in, resp., energy and application requirements failure) and closely match the optimum.

\section{Related Work}
\label{sec:relwork}

Dynamic DNNs have gained popularity for adding flexibility to  inference processes  %
and enhancing both the accuracy and efficiency of deep learning-based object detection and image classification~\cite{Trident2019,hydranets}. %
However, {\bf prior work has considered input data from only one source or the fixed deployment on a single computing node.}  In our work, we explore the use of multiple data sources and the dynamic deployment of the stem-branch architecture across the network nodes in an energy-efficient manner.

Inference of DNNs using collaborative mobile device-cloud computation has been modeled in  \cite{jointdnn2021}  using a directed acyclic graph. %
However, \cite{jointdnn2021} does not consider the dynamic variation of system parameters, multiple input sources, the stem-branch architecture, or the reliability of the inference process. 
While~\cite{malawade2022hydrafusion} delves into sensor fusion, it predominantly focuses on the training phase, neglecting the inference stage. On the other hand, \cite{EcoFusion} investigates energy-efficient inference processes but does so by selecting all available inputs, resulting in suboptimal energy utilization.
Energy is instead considered in \cite{fin_infocom2024}, which    develops a  framework to deploy  DNNs with early-exits in distributed networks.  

As for graph-based modeling  of real applications, the dynamic graph model-based optimization~\cite{HARARY199779} has been used,e.g., for wireless vehicular networks~\cite{dynamic_graph_1}. 
However, dynamic graph modeling for real-time applications requires multi-constrained temporal path discovery. In~\cite{mctp}, this has been solved using adaptive Monte Carlo Tree Search algorithm, MCTP. In our work, we compare the performance of our proposed framework to this state-of-the-art algorithm.

\section{System Scenario}
\label{sec:description}

This section first introduces the system scenario we consider by characterizing the data sources and the mobile-edge continuum  nodes, as well as their interaction. Then it describes the  dynamic gated DNN model that we developed and that we take as reference neural network model for our study.

\subsection{System components}
\begin{figure}[tb] 
\centering
  \subfigure[Sample scenario \label{f:sample_scen}]{
      \includegraphics[width=0.45\columnwidth]{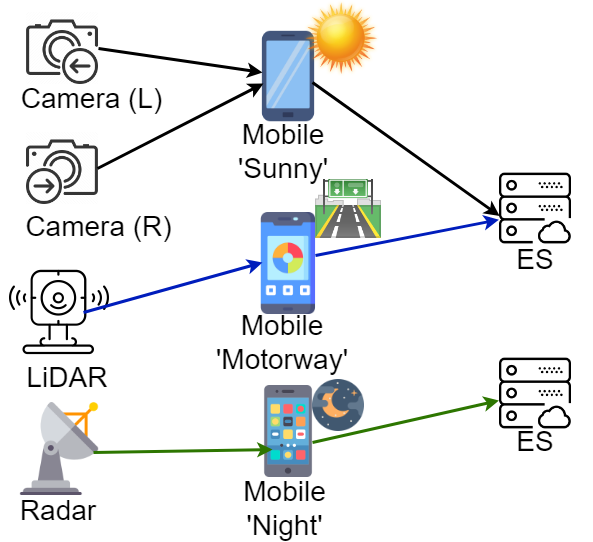}  
    }\hspace{-3mm}
         \subfigure[Static Inference Graph\label{f:sig}]{
         \includegraphics[width=0.4\columnwidth]{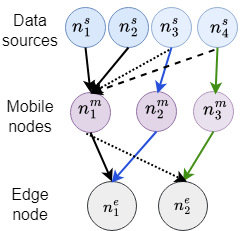}
         }
\vspace{-2mm}
    \caption{(a) System scenario with 3 applications, 4 data sources, 3 mobile nodes, each associated with a different context, and 2 edge servers. (b) Graph-based model of the system configuration. 
\label{sig_example}
}
\vspace{-0.4cm}%
\end{figure}

We consider a network system consisting of data sources (Camera (Left/Right), Radar, LiDAR), mobile devices, and edge servers.
As discussed in detail later in the paper, the notion of {\em context} is instrumental for the overall optimization, and the DNN configuration is greatly influenced by environmental conditions. Each mobile device is associated with an environmental context (e.g., \texttt{Sunny}, \texttt{Motorway}, and \texttt{Night}), which, as discussed later, drives performance given the configuration. Intelligent applications (denoted in Fig.\,\ref{f:sample_scen} using different arrow colors) require the inference task to be performed with the required level of performance guarantees (latency and accuracy). In the example scenario, Application 1 (black arrows) uses Camera (L-R) input on a mobile node in a \texttt{Sunny} context and the dynamic DNN for this application is deployed on the mobile node and the edge server (ES). Application 2 (blue arrows) uses LiDAR input and the dynamic DNN is deployed on the mobile node with the \texttt{Motorway} context and on the ES. The same ES is used to deploy both Applications 1 and 2. Application 3 (green arrows) uses Radar input and the DNN model is deployed on the mobile node with \texttt{Night} context and on a different ES. 

The devices acting as data sources, the mobile nodes, and the ESs are connected over a wireless network (i.e., data sources could be either co-located with mobile nodes or connected to them). The communication resources of the network and the computation resources of the mobile nodes and ESs are shared for executing the inference tasks using the dynamic deployment of a pre-trained dynamic DNN model.
To effectively support the applications, an orchestrator, located in the network infrastructure, (i) determines the data sources to be used to feed the DNN for the different applications, and, accordingly,  instructs the mobile nodes about the data to be collected, (ii) instructs both the mobile nodes and the edge servers about the DNN's sections  (stems or branches) to deploy locally, and generates the configuration to be used by the gates  controlling the data processing throughout the DNN. 

\begin{figure}[tb]
\centering
\includegraphics[width=0.4\textwidth, trim = 0cm 5.5cm 3cm 0cm, clip]{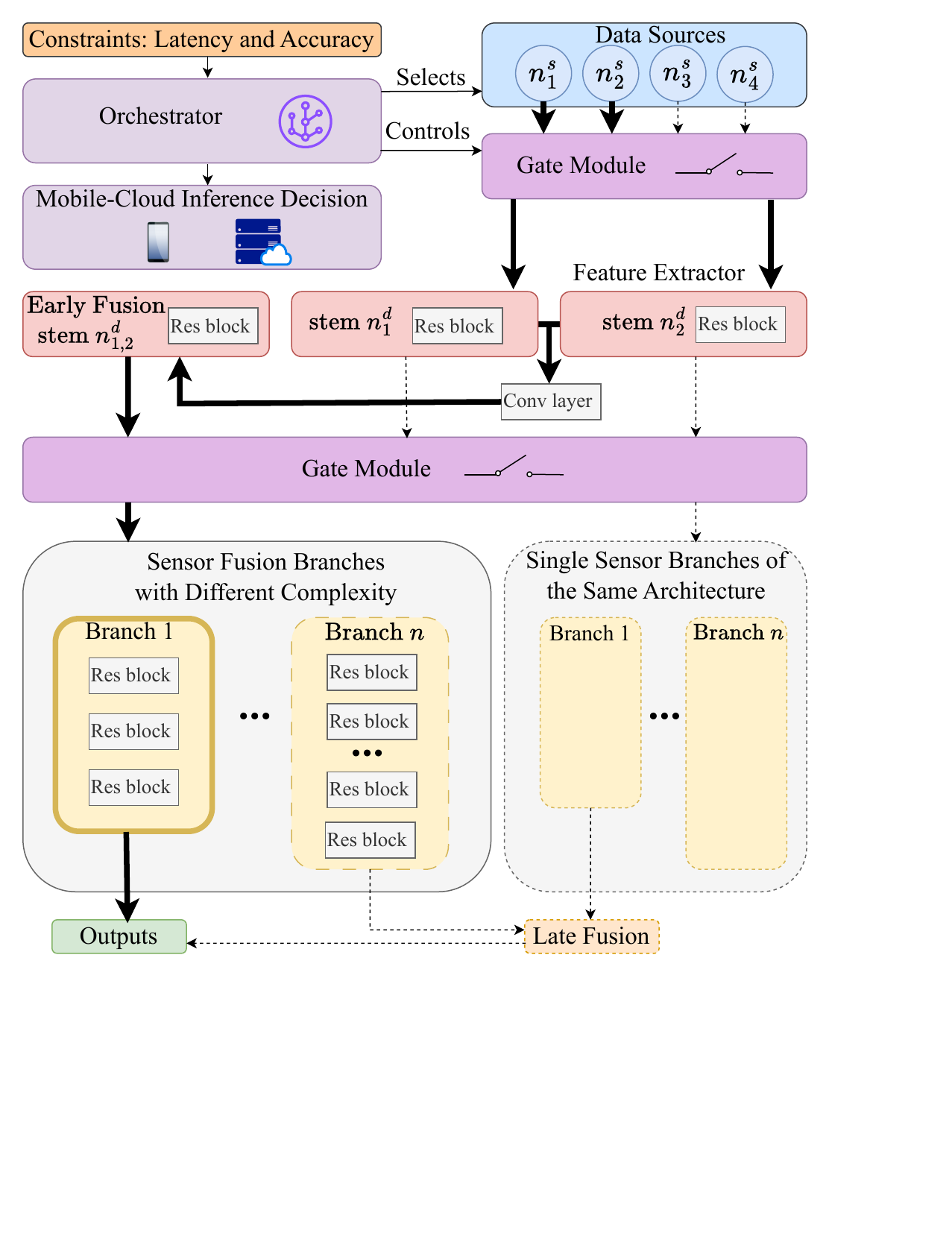}     \vspace{-4mm}\caption{\vspace{-0mm}Architecture implemented for object detection inference, illustrating the process where the orchestrator selects the data source, followed by the gate modules  determining the appropriate stem and branch for deployment.}
\label{f:architecture_example}
\vspace{-6mm}
\end{figure}

\subsection{Dynamic DNN model\label{sub:DNN_architecture}}

One of the core contributions of this paper is accounting for the deep interplay between innovative dynamic neural models and the resource allocation strategy of the collaborative edge computing system described previously.
In this section, we present the instance of dynamic gated neural network model that we developed. At a high level, the architecture performs dynamic, adaptive and context-aware sensor fusion on Camera, LiDAR, and Radar data for object detection. Fig.\,\ref{f:architecture_example} summarizes the structure of the model. Our model architecture adopts and extends the HydraFusion framework \cite{malawade2022hydrafusion}, introducing a scalable design to accommodate the varying computational demands of the applications to be supported via our gated mechanism. We remark how the adaptation of the ability of the model to change the computing workload is instrumental to integrate the models in a system-wide resource allocation framework. 

As illustrated in Fig.\,\ref{f:architecture_example}, a gate module controls the activation of the stems, determining how features from different input sensors are fused and analyzed. Contrary to the gating mechanism employed in \cite{malawade2022hydrafusion}, which selects branches during the training phase to enhance accuracy, {\bf our study implements the gating mechanism during the inference phase,} in order to optimize the tradeoff between output accuracy, energy expense, and latency given the current context. 

Our neural model architecture processes input data from various modalities to facilitate object detection. We integrate   ResNet-18/50/101\,\cite{he2015deep}, our  base architecture, within a faster Recursive-Convolutional Neural Network (R-CNN).  
Specifically, initial sensor data from various modalities are analyzed by distinct CNNs, referred to as ``stems''. These stems, serving as feature extractors, are specialized to process the respective sensor inputs, transforming them into initial sets of features. These stems correspond to the first block of the ResNet architecture. An early fusion mechanism concatenates the features from each stem, resulting in three augmented stems that enhance context-specific accuracy. A 2D convolution layer merges these concatenated features, which are inputs to the subsequent ResNet layers, termed ``branches''.  

The architecture consists of single-sensor branches as Left Camera, Right Camera, LiDAR, and Radar, alongside early fusion branches combining Left and Right Cameras, LiDAR and Radar, and Left Camera and LiDAR. This configuration results in 6 branches, each presenting 3 levels of complexity aligned with the ResNet 18/50/101 models, resulting in a total of 18 selectable branches. A second  gate module then determines the most suitable branch or branches for the task.

Furthermore, we integrate late fusion as a post-processing step exclusively for the (less complex) ResNet18 branches, constrained by computational efficiency. 
We applied  Non-Maximum Suppression (NMS) as our late fusion mechanism to the outputs from the ResNet-18 branches to balance deployment efficiency. Indeed, empirical evidence from our pre-trained models indicates that late fusion improves accuracy. However, it necessitates the complete deployment of multiple branches, which escalates energy consumption. Consequently, our gating mechanism consistently avoids selecting late fusion  for ResNet-50 and ResNet-101 branches. 

Fig.~\ref{f:architecture_example} illustrates the decision process, highlighted by bold arrows, starting from the selection of data sources, leading to the choice of early fusion branches of a specific complexity level, excluding late fusion. Dashed lines represent potential, yet unselected, pathways available to the gating module.

We remark  how different configurations of the gates result in a different computing load associated with stems and branches, as well as a different data flow input-to-stems and stems-to-branches. Further, this interrelation  heavily depends on the context. Intuitively, these quantities are quintessential to define resource allocation. We also underline how the model is splittable, i.e., stems and branches can be executed on different nodes. This results in a data flow and computing load between mobile nodes and  edge servers that is dependent on the configuration and on the allocation of computing tasks.

\section{System Model and Problem Formulation}
\label{sec:model}

We represent the above system scenario by defining a set of network nodes, $\Nc{=}\{\Nc^\text{s}{\cup}\Nc^\text{m}{\cup}\Nc^\text{e}\}$, including:
    1) data sources (hence, data modes) in~$\Nc^\text{s}$;
    2) mobile nodes (which can host either only the stems or the stems along with the branches of the DNN model) in~$\Nc^\text{m}$;
 3) edge servers (ESs) (which can host either only the branches or the stems along with the branches of the DNN model),  in~$\Nc^\text{e}$.
$N{=}|\Nc|$, $N^\text{s}$, $N^\text{m}$, and $N^\text{e}$ denote (resp.) the number of all nodes,  sources, mobile nodes, and edge servers.

For each node~$n{\in}\Nc^\text{s}{\cup}\Nc^\text{m}$, the amount of (uplink)  communication resources at its disposal, e.g., the number of resource blocks a mobile node can use to communicate with the edge servers, is denoted with ~$B_n$. 
Similarly, $C_n$ denotes the amount of computational (e.g., CPU or GPU) resources (in number of executable instructions/s) available at node~$n{\in}\Nc^\text{m}{\cup}\Nc^\text{e}$, which can be used for sensor fusion and inference.

{Applications} are denoted by elements~$h{\in}\Hc$, and are associated with a maximum latency requirement~$\ell_{\omega,\max}^h$ and a minimum accuracy requirement~$\alpha_{\omega,\min}^h$, 
{\em both expressed in terms of the quantile}~$\omega\in [0,1]$. Considering different quantiles allows us to balance efficiency (e.g., energy consumption) against inference quality and latency guarantees. We can thus limit the extent of an undesirable event (failing to meet the application requirements) by tweaking $\omega$, depending upon the scenario and application at hand. Intuitively, one might use more lightweight models in scenarios/applications where occasional failures can be accepted. On the other hand, critical scenarios where accuracy guarantees are required call for more robust models -- even if they have more substantial resource requirements. We remark how the resource usage versus accuracy performance tradeoff is informed by the context.

For each application~$h$, we have a set of possible data sources $\Nc^h{\subseteq}\Nc^\text{d}$,  corresponding to nodes equipped with a sensor (e.g., Camera, LiDAR, or Radar) that can be used for application~$h$. The quantity~$\delta_n^h$ represents the quantity of data (in bits) emitted by source~$n$ when used for  application~$h$.

Also, each application is associated with a {\em splittable} dynamic DNN model, composed of stems~$\Ac^{\text{s},h}$ and branches~$\Ac^{\text{b},h}$ (although simply referred to as branches, the latter ones may come with an additional stem at the end, as described in Sec.\ref{sub:DNN_architecture}). More specifically, the model can be split by deploying stems and branches at different nodes. For each stem and branch in~$a{\in}\Ac^{\text{s},h}{\cup}\Ac^{\text{b},h}$, and for each node~$i{\in}\Nc^\text{m}{\cup}\Nc^\text{e}$, we know the quantity $\delta^h_a$ outgoing from stem (or branch) $a$ when used by application~$h$. For each stem~$a{\in}\Ac^{\text{},h}$ (branch~$a{\in}\Ac^{\text{b},h}$) and application~$h$, we are also given the computational complexity of each stem (branch), expressed in number of operations~$o^h_a$ (e.g., in CPU cycles per bit of incoming data) necessary to run the stem (branch).

{\bf Problem formulation.} 
The decisions the orchestrator needs to make are: 
1) the data source(s), stem(s), and branch(es)  to use for each application;
    2) where to deploy them;
3) how to distribute the computation and communication resources available at nodes across applications.
Once it makes the decisions (1) and (2), then the  orchestrator also generates the logic to be executed by the gates of the dynamic neural model.  
We express the first two decisions through variables $\sigma(h,n){\in}\Ac^{\text{s},h}{\cup} \Ac^{\text{b},h}{\cup}\{\textsf{data},\emptyset\}$, expressing how application~$h{\in}$ uses node~$n{\in}$. Such variables can take the following values:
 $\emptyset$, if node~$n$ is not used by application~$h$;
    $\textsf{data}$, if that node is a data source  used by application~$h$;
   a value in $\Ac^{\text{s},h}$ or $\Ac^{\text{b},h}$, if application~$h$ uses node~$n$ to deploy a stem or branch (respectively).
We also indicate with~$\bsigma$ (in bold) the $|\Hc|\times|\Nc|$ matrix collecting the values of all $\sigma$-variables.

Concerning resource allocation, we indicate with~$c_n^h$ (in no. instructions/s) and $b_n^h$ (in no. of resource blocks), respectively, the computational and radio resources   that are allocated to application~$h$ at node~$n$, and with~$\rho_{n}$ the per-resource block bit rate associated with the highest Modulation and Coding Scheme (MCS) that node $n$ can use for uplink transmissions. Accordingly, $R_{n}^h$ denotes the (outgoing) data rate available to application~$h$ at node $n$, i.e.,  $R_{n}^h{=}b_n{\cdot}\rho_{n}$.

Given the values of the above decision variables,   the system performance can be derived as follows. The overall system energy consumption is driven by the computational and communication resources allocated at each node, hence: 
\begin{equation}
\epsilon(\bsigma,b,c){=}\sum_{n\in\Nc}\sum_{h\in\Hc}\left(\epsilon^c_n c_n^h+\epsilon^b_n b_n^h\right),
\end{equation}
where~$\epsilon^c_n$ and $\epsilon^b_n$ represent the energy consumption associated with the usage of each unit of (resp.) computational and communication resources  at node~$n$.

The time for sensor fusion and inference then includes two components: the computational latency and the network latency. Given application $h$, the former is given by:
\begin{equation}
\ell_\text{c}^h(\bsigma,c){=}\sum_{n{\in}\Nc\colon\sigma(h,n){\in}\Ac^{\text{s},h}{\cup}\Ac^{\text{b},h}}\frac{o^h_{\sigma(h,n)}}{c_n^h},
\end{equation}
while the latter is given by:
\begin{equation}
\ell_\text{b}^h(\bsigma,b){=}\text{\hspace{-5mm}}\sum_{n\in\Nc\colon \sigma(h,n)=\textsf{data}}\frac{\delta_n^h}{R_n^h}{+}\text{\hspace{-3mm}}\sum_{n\in\Nc\colon \sigma(h,n)\in\Ac^{\text{s},h}\cup\Ac^{\text{b},h}}\text{\hspace{-3mm}}\frac{\delta^h_{\sigma(h,n)}}{R_n^h}.
\end{equation}
In other words, computational latency is incurred for each node that is used for either a stem or a branch (top), while network latency (bottom)  accounts for the size of data and available data rate. Combining the above, the total  latency for application $h$ is:
$\ell^h(\bsigma,b,c)=\ell_\text{c}^h(\bsigma,c)+\ell_\text{b}^h(\bsigma,b)$,
and we indicate with $\ell_\omega^h(\bsigma,b,c)$ the $\omega$-quantile of the sensor fusion and inference time and with $\omega\in[0,1]$ the target {\em quantile} we are interested into. 
Notice that, in general, $\ell_\omega^h(\bsigma,b,c)$  depends upon the distribution of the bitrate resulting from the selected MCS. 
Concerning accuracy, 
its distribution and quantiles 
can be estimated through several methodologies, e.g.,~\cite{malawade2022hydrafusion}, 
yielding the accuracy {\em quantile}~$\alpha_\omega^h(\bsigma)$.

Considering a {\bf snapshot of the system under study  modeled through the static sensor fusion and inference graph}, and combining all the above, we can formulate  the orchestrator's decision problem as the following optimization problem: 
\begin{subequations} \label{opt_total}
\vspace{-3ex}
\begin{align}
& \min_{\bsigma,b,c}\epsilon(\bsigma,b,c) & \label{eq:obj}\\
\text{s.t. } & \alpha_\omega^h(\bsigma) \geq \alpha^h_{\omega,\min}  & \forall h\in\Hc \label{eq:constr-alpha}\\
& \ell_\omega^h(\bsigma,b,c)  \leq\ell^h_{\omega,\max}   & \forall h\in\Hc \label{eq:constr-ell}\\
& \sum_{h\in\Hc} c_n^h\leq C_n &  \forall n\in\Nc \label{eq:constr-c}\\
& \sum_{h\in\Hc} b_n^h\leq B_n &  \forall n\in\Nc \label{eq:constr-b}.
\end{align}
\end{subequations}
The orchestrator seeks to minimize the total energy consumption \Eq{obj}, subject to the constraints that all applications achieve their target accuracy quantile \Eq{constr-alpha} within their target application latency \Eq{constr-ell}. In doing so, the orchestrator must be mindful of the total amount of computational \Eq{constr-c} and radio \Eq{constr-b} resources available at each node. 
Intuitively, obtaining a better performance (thus satisfying \Eq{constr-alpha} and \Eq{constr-ell}) tends to require more resources, but doing so would increase the energy consumption \Eq{obj}. At the same time, \Eq{constr-c} and \Eq{constr-b} prevent the orchestrator from using particularly desirable (e.g., well-connected) nodes beyond their capabilities.

\begin{property}
The problem of minimizing \Eq{obj} subject to constraints \Eq{constr-alpha}--\Eq{constr-b} is NP-hard.
\end{property}
\begin{IEEEproof}
We prove the property via a reduction from the knapsack problem, which is NP-hard.%
Given an instance of the knapsack problem with  a set a set~$\mathcal{I}{=}\{i\}$ of items with weight~$w_i$ and value~$v_i$, and given the maximum weight~$w^{\max}$, we can build  a {\em heavily simplified}, one-application version of our problem where items are mapped into data sources, the source corresponding to item~$i$ increases the energy consumption by~$v_i$ and the accuracy by~$w_i$, the accuracy target is equal to~$w^{\max}$, and selecting an item is mapped into {\em not} selecting the corresponding data source. Then solving our problem %
is equivalent to solving the knapsack problem. %
Finally, it can be seen by inspection that reduction is polynomial in complexity -- indeed, constant, as it has no loops.

\end{IEEEproof} 

In summary, although helpful to formalize the problem that the orchestrator needs to face,  {\em the problem complexity and the fact that the  mobile  network system and context 
are dynamic in nature} (they  both vary with time) demand for an efficient algorithmic  solution strategy.  Below, we address this need by proposing our QIC  solution, which effectively and efficiently  copes with both the system complexity and dynamics.

\section{QIC: A Dynamic Dependable Solution}
\label{sec:solution}
Given the dynamic nature of the system, in our solution approach we introduce the notion of time by extending the static sensor fusion and inference graph model 
(discussed in the previous section) to an {\em attributed dynamic graph model}. Then, in light of the problem complexity, we also apply the {\em   Quantile Constrained Policy Optimization (QCPO) reinforcement learning algorithm~\cite{qcpo} to the attributed dynamic graph  and find the efficient set of data sources to be used and the DNN deployment configuration and resource allocation} in the dynamic system for heterogeneous applications.

\begin{figure}[t] 
\centering
\includegraphics[width=0.3\textwidth]{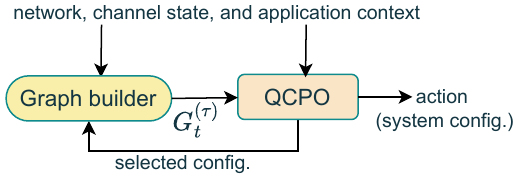}
\vspace{-4mm}
\caption{QIC solution framework.}
\label{qcpo_approach}
\vspace{-4mm}
\end{figure}

A schematic illustration of our proposed QIC  framework  is presented in Fig.\,\ref{qcpo_approach}. 
It consists of two blocks: {\em the Graph Builder} and the {\em QCPO}. Given time $t{=}0$, the first block  creates the initial attributed dynamic graph, $G_t^{(0)}$, i.e., the attributed graph reflecting the system at the initial time instant. The QCPO block instead performs quantile constrained reinforcement learning (RL) \cite{qcpo}.  It  takes $G_t^{(0)}$ as input and selects the action, i.e., the  configuration (DNN stems and branches, where they are deployed and the corresponding resource allocation), that maximizes a reward function matching the objective  in (\ref{eq:obj}). Based on the selected action,  the Graph Builder updates  the attributed dynamic graph,  
yielding $G_t^{(1)}$.  The procedure is repeated for  $P_{\max}$ epochs, thus generating $G_t^{(\tau)}$ at every epoch $\tau$, and the solution to be enacted at time $t$ will be given by the action selected in the last epoch.   

Below,  we give further details on our solution approach.

\subsection{Attributed dynamic graph model}
\begin{figure}[!tb] 
\centering
\includegraphics[width=0.49\textwidth]{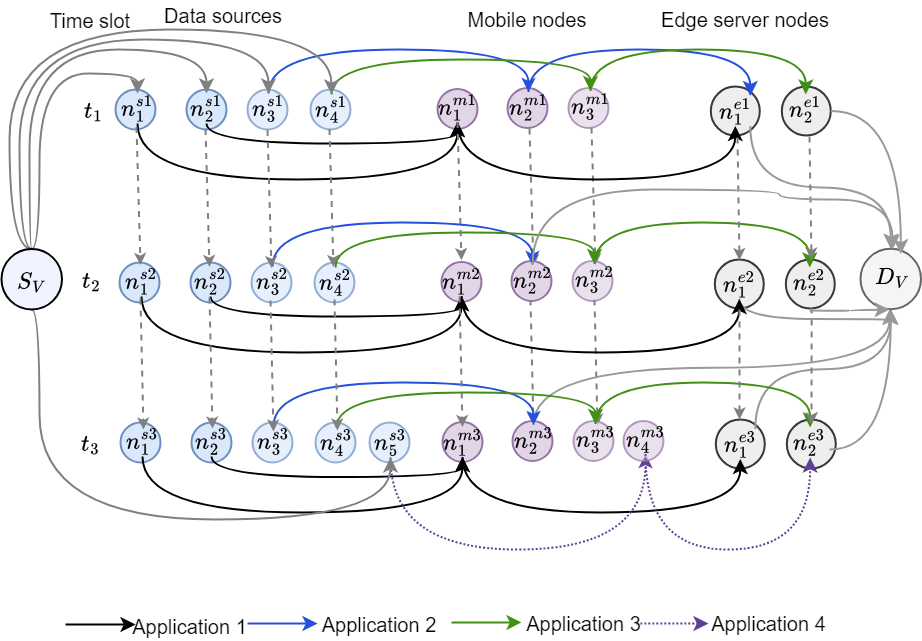}
\vspace{-5mm}
    \caption{Dynamic sensor fusion and inference graph  over three time slots. At $t_1$,  it reflects the example  in Fig.\,\ref{sig_example}; at $t_3$, it includes   4 applications, 5 data sources, 4 mobile nodes, and 2 edge servers. }%
\label{f:dynamic_graph}
    \vspace{-4mm}
\end{figure}

We account for the temporal variations of the  applications,  network conditions,  and context (e.g., \texttt{Sunny}, \texttt{Motorway}, \texttt{Night}) by  combining the static sensor fusion and inference graph representation of the system  (Fig.\,\ref{sig_example}) given for each time instant into an attributed dynamic graph model (Fig.\,\ref{f:dynamic_graph}).
By doing so, the attributed dynamic graph  can represent a time-based  deployment of dynamic DNNs with stems and branches in the mobile-edge continuum to provision sensor fusion and inference tasks for heterogeneous applications. 

Each edge in an attributed dynamic graph contains multiple dynamic attributes, each   specifying a different system-level constraint. This facilitates the identification through the graph of the solution to our time-varying, multi-constrained  problem.

The dynamic graph is  $\mathcal{G}{=}\{G_1,G_2,{\ldots} G_T\}$, where:
    
    \textbullet\, $G_t{=}\{V_t,E_t,\mathcal{F}_t\}$, $t{=}1,{\ldots}T$,  models the system at time\,$t$;
    
    \textbullet\,  $V_t$ is the set of vertices of $G_t$, representing  data sources $n^{st}_i, i\in\{1,2,\ldots N^s\}$, mobile nodes $n^{mt}_i, i\in\{1,2,\ldots N^m\}$, and edge servers $n^{et}_i$, $i{\in}\{1,2,\ldots N^e\}$,  plus a source, $S_v$, and a destination,  $D_v$, as fictitious vertices representing (resp.) the starting and ending point of the system configuration process;
    
    \textbullet\, $E_t$ is the set of edges of $G_t$.  An edge $e{\in} E_t$ is defined as a tuple $\left( u, v, \{ b^h_u, c^h_u\}_{h{\in} [1,H]}, \rho_u\right)$ where: (i) $u,v {\in} V_t$, (ii) the number of resource blocks ($b_u^h$) and compute resources ($c_u^h$) is the   assignment for each application $h$ at node $u$ such that $\sigma(h,u){=}a$, with $a{\in}\Ac^{\text{s},h}\cup\Ac^{\text{b},h}$, if one or more stems or branches of the DNN  are deployed on node $u$ for application $h$, (iii) and $\rho_u$ is the uplink per-resource block data rate at $u$. The same holds if $\sigma(h,u){=}\textsf{data}$, but for the compute resources allocation which in this case is $c_u^h{=}0$;
    
    \textbullet\, By setting $K$ to the number of constraints plus the objective function in our optimization problem  \eqref{eq:constr-alpha}--\eqref{eq:constr-b}  (i.e.,  $K{=}5$), ${\mathcal{F}}_t{=}\{f_1, \ldots f_K\}$ is the set of  normalized constraint attribute functions
that assign to every edge  $e{\in}E_t$ a non-negative value $f_j(e)$, with  $f_j(e){:}\left(u,v,\{\
b^h_u,c^h_u\}_h, \rho_u\right)$ ${\in} E_t{\to}\mathbb{R}^+$, $j{\in}\{1,\ldots K\}$.  Each $f_j(e)$ corresponds to the value of the objective function (if $j{=}1$) or of a normalized constraint (if $j{=}2,\ldots K$), setting the right hand side in the constraints expression to 1, associated with edge $e$.

\noindent
{\bf Computational complexity.} 
Generating the dynamic graph has low, namely, polynomial computational complexity. Indeed, the graph has at most $N{+}2{=}N^s{+}N^m{+}N^e{+}2$~vertices, hence,  $O(N^2)$~edges. Each edge has $2H{+}1$~attributes, giving a total complexity of~$O(N^2{+}H)$; assuming that there are more nodes than applications, the complexity can be further simplified to~$O(N^2)$, i.e., quadratic in the number of nodes.

\subsection{The QCPO solution framework}

Using the QCPO approach~\cite{qcpo}, the decision process can be modelled as a constrained Markov decision process and, accordingly, the dynamic system can be   defined by the tuple ${<}\mathcal{S}^s,\mathcal{S}^a, \textbf{r},\textbf{c},{\mathbf M}, \gamma{>}$, where $\mathcal{S}^s$ is the state space, $\mathcal{S}^a$ is the action space, $\textbf{r}:\mathcal{S}^s{\times}\mathcal{S}^a{\rightarrow}\mathbb{R}$ is the reward function, $\textbf{c}:\mathcal{S}^s{\times}\mathcal{S}^a{\rightarrow}\mathbb{R}^+$ is the cost function, ${\mathbf M} :\mathcal{S}^s{\times}\mathcal{S}^a{\times}\mathcal{S}^s{\rightarrow}[0,1]$ is the state transition probability, and $\gamma$ is a discount factor used for computing the accumulated cost of the Markov decision process over the epochs.  
In the following, we fix the time instant $t$ and drop the dependency on $t$ whenever clear from the context. We instead denote with $\tau$ the  epoch of the QCPO process, which is performed at each $t$ to adapt the selected configuration to the system's dynamics.  
The system state is then given by the set $\textbf{s}_t{=}\textbf{s}_{\tau}{=}\{B_n, C_n,\rho_n\}_n$ and an action is represented by the set $\textbf{a}_{\tau}{=}\{\bsigma,b^h_n,c^h_n\}_{h,n}$ (with $\textbf{a}_t{=}\textbf{a}_{P_{\max}}$, i.e.,   the  action to be enacted at time $t$ is the one selected at $\tau{=}P_{\max}$). 
QCPO implements a policy $\pi$, which selects the action maximizing the reward function (specified below).  The edges ($E_t$) and attribute function values ($\mathcal{F}_t$) of the dynamic graph $G_t^{(\tau)}$  are updated at every epoch ($\tau{\in}[1,P_{\max}]$) based on system state %
and action. 

We define the reward function as a positive (negative) inverse of energy consumption based on success (failure) in meeting the system and application constraints, given by:
\begin{align}
&\textbf{r}(\textbf{s}_\tau,\textbf{a}_\tau)=\sum\limits_{e\in E_t}\frac{\psi(e)}{1+f_1(e)}\label{eq:reward}\\
\psi(e)=&\begin{cases}
			1, & \text{if $f_j(e)\leq1,j{=}2,\ldots K$, i.e.,  \eqref{eq:constr-alpha}--\eqref{eq:constr-b} are met}\\
            -1, & \text{otherwise.}
		 \end{cases}\nonumber
\end{align}
We also define the cost function as the weighted (weight $\mu_j$) sum of the normalized constraint attribute functions  that correspond to the problem constraints, i.e., 
$\textbf{c}(\textbf{s}_\tau,\textbf{a}_\tau)=\sum\limits_{e\in E_t}\sum\limits_{j=2}^{5}\mu_jf_j(e)$. 
The estimated cumulative sum cost is:
$X^\pi(\textbf{s}){=}\sum\limits_{\tau{=}0}^\infty\gamma^\tau\textbf{c}(\textbf{S}_\tau,\textbf{A}_\tau)$ 
where, given $\pi$  the policy function, $\textbf{A}_\tau {\sim} \pi(\cdot|\textbf{S}_\tau)$ and $\textbf{S}_{\tau}{\sim} {\mathbf M}(\cdot|\textbf{S}_{\tau-1},\textbf{A}_{\tau-1})$.
Importantly, the policy $\pi$ implemented by QCPO   selects an action so that the quantile of the distribution of the $X^\pi(\textbf{s})$ does not exceed a specified threshold ($d_{th}$). By doing so, it satisfies the system latency and accuracy constraints within the selected quantile.  

QCPO implements a policy $\pi$ maximizing the  reward and limiting the quantile of the cumulative cost by leveraging two neural networks, namely, the policy and value neural networks. The former, also known as
actor, is used to choose actions and update the policy; the latter,  also known as critic,  is used to estimate the value
function. 
More specifically, the original optimization problem, given in \eqref{opt_total}, is modified for QCPO and expressed in terms of expected cumulative reward  and quantile of the estimated cumulative cost  functions, as follows:
  \begin{subequations} \label{opt_qcpo}
  \begin{align}
     \quad \max_{\pi}\,\, &   V^\pi(\textbf{s}_0)=\mathbb{E}_\pi\left[\sum_{\tau=0}^\infty\gamma^\tau\mathbf{r}(\textbf{s}_\tau,\textbf{a}_\tau)\right]\label{qcpo0}\\
 \text{s.t.} & \quad
     q^\pi_{\omega}\leq d_{th}
     \label{qcpo1},
    \end{align}
\end{subequations}
where 
 $\omega$-quantile of the r.v.\,of the estimated cumulative sum cost is $q^\pi_\omega(\textbf{s}){=}\inf\{x|\text{Pr}(X^\pi(\textbf{s}){\leq} x) {\geq} \omega\}$. 
We compute the quantile and tail probability of the  cumulative sum cost (modelled through the Weibull distribution) via the distributional RL with the large deviation principle~\cite{qcpo}. %
QCPO meets the 
 constraint after the  RL policy and value network training.

\textbf{Policy and value networks.} 
The  objective function of the policy network  is based on the parameter $\phi$ given by:
$L(\boldsymbol{\phi}) {=}\textbf{E}[{J(\boldsymbol{\phi})}]$. 
 The  objective function of the value network is:
$L(\boldsymbol{\Omega}) =\textbf{E}[{J(\boldsymbol{\Omega})}]$, 
where, ${J(\boldsymbol{\Omega})}$ represents the temporal difference error of the value function.

QCPO uses the proximal policy optimization (PPO)~\cite{ppo} to enable  frequent policy optimization depending on earlier policy and, in general, it  shares the parameters between policy and value networks. 
By performing clipping operation on the policy probability ratio (PPR), i.e., $\textbf{p}(\boldsymbol{\phi}){=} \frac{\pi_{\phi}(\textbf{a}_\tau|\textbf{s}_\tau)}{\pi_{\phi_{o}}(\textbf{a}_\tau|\textbf{s}_\tau)}$, the clipped policy objective function  
is:
$J(\boldsymbol{\phi})=\mathbb{E}\Bigl{[} \mathcal{\hat{L}} (\textbf{p}(\boldsymbol{\phi})) \hat{A}_{\pi_{\boldsymbol{\phi}_{o}}}(\textbf{s}_\tau,\textbf{a}_\tau)\Bigl{]}$ 
where: 
$\mathcal{\hat{L}}(\textbf{p}(\boldsymbol{\phi}))$ is 
$ 1-\theta %
 \text{ for } \textbf{p}(\boldsymbol{\phi})\leq 1-\theta;      
 1+\theta \text{ for }  \textbf{p}(\boldsymbol{\phi}) \geq 1+\theta; \text{ and }  
  \textbf{p}(\boldsymbol{\phi})$   \text{otherwise.}
Here, $\pi_{\boldsymbol{\phi}_{o}}(\textbf{a}_\tau|\textbf{s}_\tau)$ and $\hat{A}_{\pi_{\boldsymbol{\phi}_{o}}}(\textbf{s}_\tau,\textbf{a}_\tau) $ represent the previous policy and estimate of advantage function, respectively, while  $\theta$ is the clipping parameter. $\boldsymbol{\phi}_{o}$ is the policy parameter prior to update. The value (quantile) advantage function is an estimate of the difference between the total weighted reward (cost) and the estimated value (quantile) function for a selected action, on the completion of an epoch. A positive value means that the chosen action is preferred. QCPO takes the policy gradient using the sum of the value and the quantile advantage~\cite{qcpo}. The update expression of the policy network parameter for each proposed policy is: 
\begin{equation}
\boldsymbol{\phi}_{t+1}=\boldsymbol{\phi}_t+\frac{\eta}{P_{\max}}  \sum_{\tau=1}^{P_{\max}}\bigtriangledown_{\phi}J(\boldsymbol{\phi}) 
\label{eq:policy_update}
\end{equation}
where $\bigtriangledown_{\phi}$ is the policy objective function gradient. 

The overall state-based (on-policy) procedure of QIC using the QCPO action selection function is given in Algorithm \ref{alg:qcpo}.

\vspace{-2mm}
\begin{algorithm}[!htb]
\begin{minipage}{0.9\linewidth}
\small
\caption{QCPO Algorithm}
\label{alg:qcpo}
\small
\begin{algorithmic}
\STATE {\textbf{Input:} $\textbf{s}_t$}\\ 
 \SetKwFunction{FMain}{Main}
  \SetKwFunction{FSum}{\!\!}
  \SetKwFunction{FSub}{\!\!}
\noindent\SetKwProg{Fn}{Function I: QCPO\_action\_selection}{:}{}
\Fn{\FSub{$\textbf{s}_t$}}{
  \begin{enumerate}[leftmargin=-2pt]
\item  {Init.\,policy neural network with rand.\,seed, $\boldsymbol\phi_t{\in}(0,1]$}
\item  {Init. state value function with random seed, $\boldsymbol\Omega {\in} (0,1]$};
    \FOR {$\tau=1,2,\ldots P_{\max}$ }
\STATE 3.1) Observe the present state $\textbf{s}_\tau$
 \STATE   3.2) Compute current reward $\textbf{r}(\textbf{a}_\tau,\textbf{s}_{\tau})$ using \eqref{eq:reward}
  \STATE   3.3) Set $temp=0, \textbf{a}_\tau=NULL$;
\FOR {$a \in \mathcal{S}^a$ }
\STATE 3.4) Estimate $\textbf{s}_{\tau+1}$ based on action $a$
\STATE 3.5) Compute $\textbf{r}(a,\textbf{s}_{\tau+1})$
\IF{\big{(}$\textbf{r}(a,\textbf{s}_{\tau+1})>temp$\big{)}}
\STATE {$temp=\textbf{r}(a,\textbf{s}_{\tau+1})$;}
\STATE {$\textbf{a}_\tau=a$;}
\ENDIF
   \ENDFOR
   \STATE 3.6) Estimate the value function $V^\pi(\textbf{s}_\tau)$ for return
   \STATE 3.7) Estimate the quantile function $q^\pi_\omega(\textbf{s}_\tau), \omega\in\{ \omega_1, \omega_2,\ldots \omega_{q}\}$ 
   \STATE 3.8) Approx. right tail distrib. $P_{X^\pi(\textbf{s}_\tau)}$  via Weibull distrib. and compute advantage functions~\cite{qcpo}. %
   \STATE {3.9)} Take policy gradient using sum of value and quantile advantage~\cite{qcpo}.%
   \setcounter{enumi}{2}
   \ENDFOR
   \item $\textbf{a}_t=\textbf{a}_{\tau}$
\item {Update $\boldsymbol{\phi}_{t+1}$ via \eqref{eq:policy_update}
and $\boldsymbol{\Omega}$ to maximize $L(\Omega)$}%
\end{enumerate}
  \KwRet $\textbf{a}_t$
  }
\noindent\textbf{Output:} $\textbf{a}_t$
\end{algorithmic}
\vspace{-1mm}
\end{minipage}
\end{algorithm}
{\bf Computational complexity.} 
The QCPO action selection function has polynomial computational complexity. Indeed, we have at most $P_{\max}$~epochs and $|\mathcal{S}^a|$ number of actions. Given that $|\mathcal{S}^a|$ is $O(|\Hc|\cdot|\Nc|\cdot |\Ac^{\text{s},h}|\cdot|\Ac^{\text{b},h}|)$, the total complexity of QCPO is~$O(P_{\max}\cdot|\Hc|\cdot|\Nc|\cdot |\Ac^{\text{s},h}|\cdot|\Ac^{\text{b},h}|)$.

\section{Reference Scenario}
\label{sec:refscen}

In this section, we describe the sensors dataset we use for our performance evaluation, as well the radio link  measurements we carried out to account for real-world  conditions. 

{\bf Dataset and dynamic DNN model.} We employ the RADIATE dataset \cite{sheeny2020radiate}, which offers data of Navtech CTS350-X Radar, a Velodyne HDL-32e LiDAR, and left and right ZED stereo camera for autonomous vehicle perception in diverse weather conditions such as \texttt{Sunny},  \texttt{Night}, and \texttt{Motorway}. The variety of weather conditions represents different contexts associated with different levels of difficulty in achieving a satisfactory value of inference accuracy.  Thus, this dataset presents challenges for object detection across varying environmental scenarios, prompting our investigation into dynamic model deployment tailored to distinct operational constraints.

Our dynamic DNN (see Sec.\,\ref{sub:DNN_architecture}) integrates early fusion techniques   \cite{malawade2022hydrafusion}, leveraging a ResNet-18 backbone \cite{he2015deep}. We extended it to ResNet-50 and ResNet-101, thereby offering varied complexity levels for enhanced adaptability. Furthermore, we partitioned the ResNet architecture, designating the initial block as the stem for each sensor modality, and then applied early fusion for merging these stems. The integrated  features are  then processed through the subsequent ResNet branches \cite{malawade2022hydrafusion}, and late fusion was added as a post-processing step only for the (less complex) ResNet 18 branches.

The system's architectural complexity and the size of input data  from different modalities are summarized in Tables\,\ref{t:config} and \ref{t:n_parameters}. Table\,\ref{t:config} outlines the size of the raw sensory inputs and the corresponding outputs from the initial processing stage (stem), which serve as inputs to the various branches.  Table\,\ref{t:config} also presents the computational load in terms of floating point operations per second (FLOPS). 
Table\,\ref{t:n_parameters} details the complexity of different branches, including the total number of parameters (in millions) and the FLOPS for each branch. The architectures vary from specific sensory branches, e.g., CameraBranch18, RadarBranch18, and LidarBranch18, to more complex fusion architectures, e.g.,  DualCameraFusion101 and RadarLidarFusion101. The branches denoted with '18', '50', and '101' indicate the architecture's depth. The parameter count reflects the model's size, while the FLOPS  give insight into the computational load during inference. 

\begin{table}
\caption{Size of stem raw input/output and complexity \vspace{-2mm}}
\label{t:config}
\centering
\small
\begin{tabular}{|c|c|c|c|}
\hline
\textbf{Data} & \textbf{Stem In.} & \textbf{Stem-Branch} & \textbf{FLOPS [G]} \\ \hline
L/R cam. & 672x376 & 64x168x94 & 3.552  \\ \hline
Radar polar & 1152x1152 & 64x288x288 & 31.00 \\ \hline
Lidar proj. & 672x376 & 64x168x94 & 5.900 \\ \hline
\end{tabular}
\vspace{-5mm}
\end{table}

\begin{table}[htbp]
\caption{Number of parameters and complexity of the branches\vspace{-2mm}}
\label{t:n_parameters}
\centering
\small
\begin{tabular}{|c|c|c|}
\hline
\textbf{Architecture} & \textbf{No.\,of param.\,(M)} & \textbf{FLOPS (G)} \\
\hline
CameraBranch18 & 40.20 & 21.76 \\
\hline
RadarBranch18 & 40.20 & 115.86 \\
\hline
LidarBranch18 & 40.20 & 23.00 \\
\hline
DualCameraFusion18 & 40.28 & 270.8\\
\hline
RadarLidarFusion18 & 40.28 & 586.6\\
\hline
CameraLidarFusion18 & 40.31 & 286.6\\
\hline
CameraBranch50 & 165.06 & 85.14 \\
\hline
RadarBranch50 & 165.06 & 352.5\\
\hline
LidarBranch50 & 165.06 & 89.10\\
\hline
DualCameraFusion50 & 165.06 & 982.6\\
\hline
RadarLidarFusion50 & 165.06 & 2202\\
\hline
CameraLidarFusion50 & 165.06 & 1084\\
\hline
CameraBranch101 & 184.05 & 184.1\\
\hline
RadarBranch101 & 184.05 & 573.4\\
\hline
LidarBranch101 & 184.05 & 132.4 \\
\hline
DualCameraFusion101 & 184.05 & 1496\\
\hline
RadarLidarFusion101 & 184.05 & 3434\\
\hline
CameraLidarFusion101 & 184.05 & 1562\\
\hline
\end{tabular}
\vspace{-4mm}
\end{table}

{\bf Radio link measurements.}  
As  radio link performance is key to 
 establish a balance between edge computational and communication demands,  
we incorporate real-world radio link  traces into the evaluation of  QIC. Through real-world experiments, we collected  TCP throughput and MCS values under two contexts: Outdoor and Indoor.  The former  is an outdoor open space with low interference from other devices and objects, the latter is an environment with walls as obstacles. 
We used a mobile node that travels a fixed trajectory while transmitting data through simultaneous  connections through WiFi (802.11ac) for 150\,s. Our mobile node is composed of a 4-wheel outdoor rover equipped with an Nvidia Jetson Nano operable through Linux OS and a WiFi antenna dongle to transmit data.  
In the Outdoor context, %
the mobile node follows a circle trajectory with a radius of ${\sim} 6$\,m. This circular  trajectory facilitates the variation in the signal strength due to the changes in the distance between the receiver and the transmitter. 
In the Indoor case, instead, the  node follows an L-trajectory so that, at the turn, multiple walls are between the transmitter and receiver. 
Each experiment, lasting 150\,s, is characterized by a number of mobile node's TCP connections ranging from 2 to 128. 
TCP packets are generated using the iPerf3 tool, and we capture the link performance every 100\,ms with \textit{iw} \cite{LinuxWirelessIW} link status tool available in Linux OS.  

\begin{figure}
    \centering
      \subfigure[\empty\label{f:bandwidtha}]{
      \includegraphics[width=0.45\columnwidth]{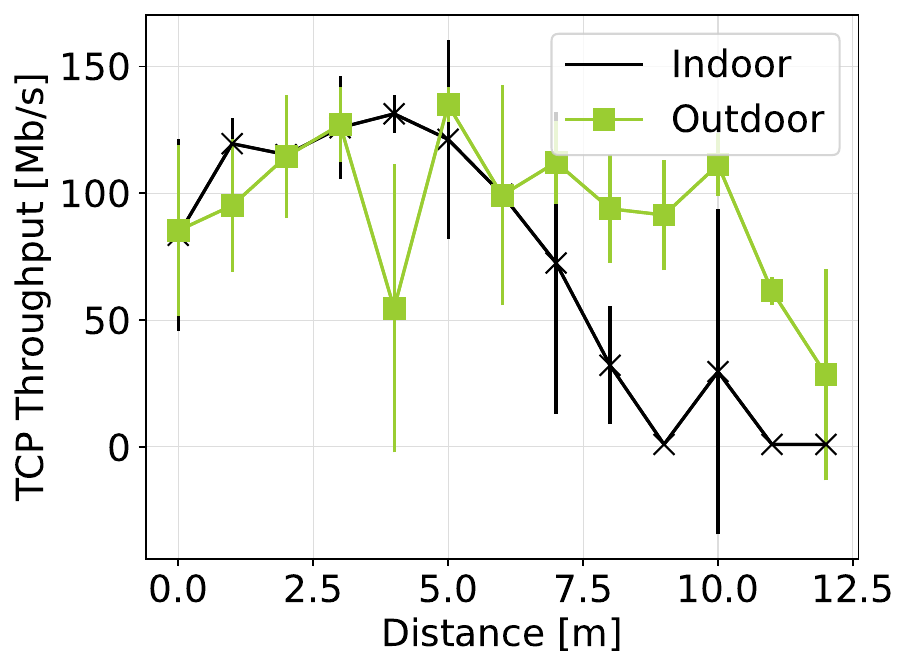}  
    }\hspace{-2mm}
         \subfigure[\empty\label{mcsc}]{
             \includegraphics[width=0.45\columnwidth]{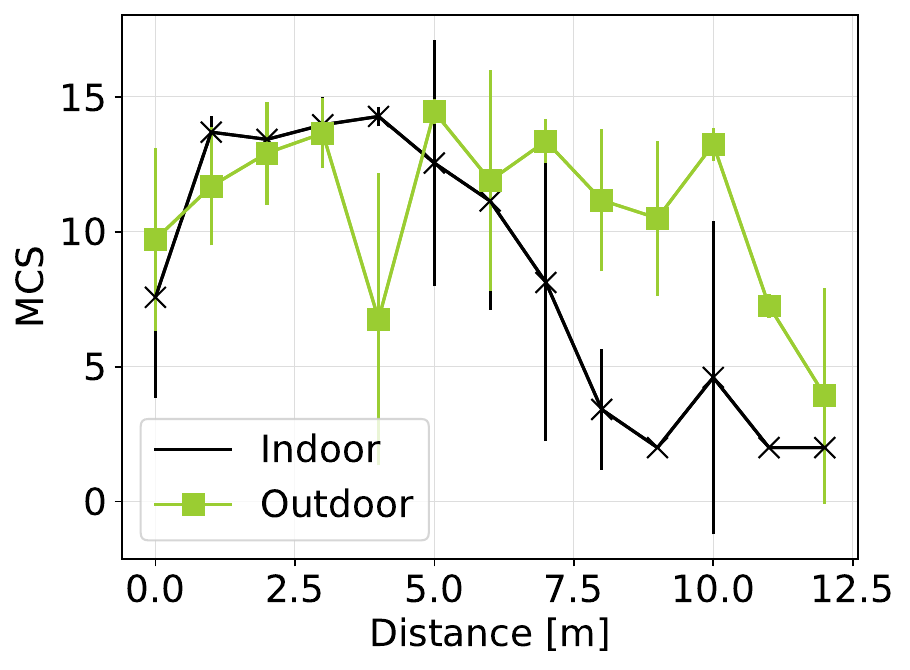}
             }
              \vspace{-3mm}
    \caption{Transport layer throughput (left) %
    and MCS (right) over the radio link, for increasing distance between transmitter and receiver.}
    \label{f:dataset}
    \vspace{-5mm}
\end{figure}

Fig.\,\ref{f:dataset}  compares the Outdoor and Indoor contexts using two metrics: transport-layer throughput (aggregated over all the mobile node's active TCP connections) and  MCS. As expected, the throughput decreases for both contexts as the distance between the ES and the mobile node grows. However,  the Indoor throughput decreases faster than the Outdoor, due to the presence of 
obstacles between  transmitter and receiver. %
In both cases, the throughput measurements are highly correlated with the  MCS  index.  
Further, Fig.\,\ref{f:dataset} shows how the variance in the MCS value for the Indoor context is higher than in the Outdoor context.  The MCS index represents the maximum data rate possible for a given channel bandwidth:  the larger the  bandwidth, the higher the data rate, hence the throughput. 

Importantly, the performed measurements give a good indication of the data transfer performance  through an OFDM-based radio interface.
For QIC's evaluation,  
we thus compute $\rho_n$  for 
the 5G NR frequency range \,1  \cite{nr12}  using the MCS index measured over time and the radio parameters   in Table~\ref{t:sim}. We also compute 
$B_n$ (the available no.\,of resource blocks) as the ratio of the measured throughput to the obtained value of\,$\rho_n$.  

\begin{table}[!tb]
\caption{Parameter settings\vspace{-4mm}}
\label{t:sim}
\begin{center}
{\footnotesize
    \begin{tabular}{|l|l||l|l|}
        \hline
       {\bf{ Parameter}}   & {\bf{Value}} & {\bf{ Parameter}}   & {\bf{Value}} \\    \hline 
{\#epochs}& {50}       &
Learning rate (QCRL)& $10^{-3}$ \\ \hline
Update parameter, $\varrho$& $0.001$ & 
{\#neurons}& {300}       \\ \hline
Update rule & Adam        & 
Steps per epoch & 10000  \\ \hline
Tests per epoch & 10  &
Steps per test & 2000  \\ \hline
 {Channel bandwidth} &  50 MHz         &
 {Carrier spacing}&15 KHz\\ \hline
 {Frequency}	&  3.4 GHz &
 {Number of data carriers} & 1200\\ \hline
\end{tabular}
}
\end{center}
\vspace{-5mm}
\end{table}

\section{Performance Evaluation}
\label{sec:peva}

We now show QIC's  performance  in a small- and a large-scale dynamic scenario, against the following  benchmarks:  

\textbullet~\emph{Multi-Constrained Shortest Temporal Path selection (MCTP)}~\cite{mctp},  applied on the  attributed  dynamic graph.  It is based on an adaptive Monte Carlo Tree Search, which  finds a path  between a graph source and destination nodes  so as to  satisfy the multiple end-to-end constraints on the attributed edge weights.  We select~\cite{mctp} because no scheme exists that specifically tackles the problem at hand.

\textbullet~\emph{Optimum (Opt),} obtained through  exhaustive search   (only in the small-scale scenario where its computation is feasible).   

\begin{figure*}
\subfigure[$\ell^h_{\omega,\max}=$50\,ms \label{f:tot_energy_sunnya}]{
\includegraphics[width=0.31\columnwidth]{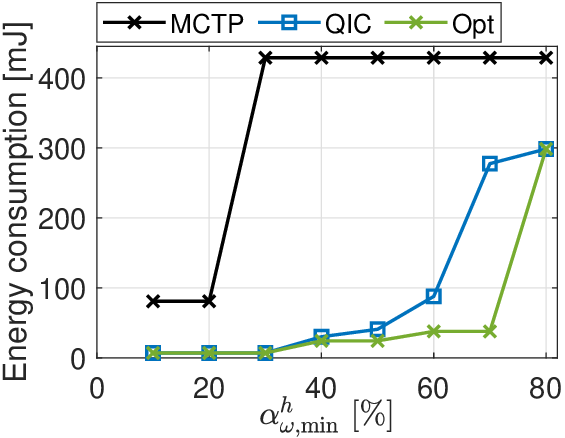}  
    }\hspace{-1mm}  \subfigure[$\alpha^h_{\omega,\min}=$50$ \%$\label{f:tot_energy_sunnyb}]{      \includegraphics[width=0.31\columnwidth]{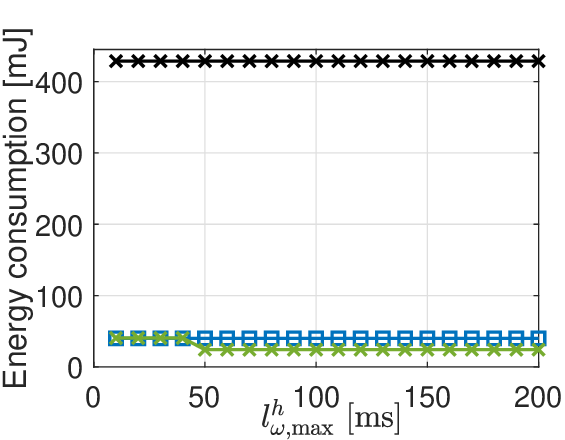}
         }
         \hspace{-1mm} 
\subfigure[$\ell^h_{\omega,\max}=$50\,ms \label{f:tot_energy_nighta}]{
\includegraphics[width=0.31\columnwidth]{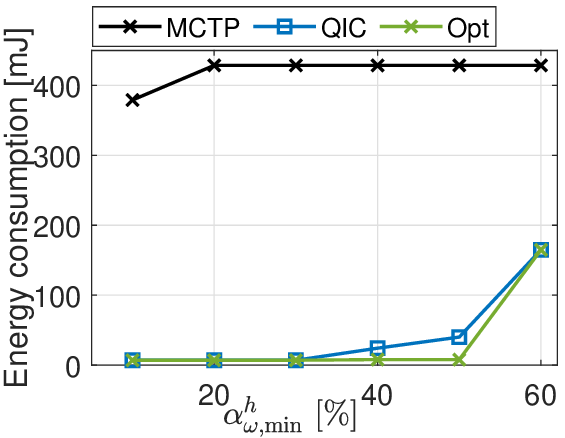}  
    }\hspace{-1mm}
\subfigure[$\alpha^h_{\omega,\min}=$50$ \%$\label{f:tot_energy_nightb}]{   \includegraphics[width=0.31\columnwidth]{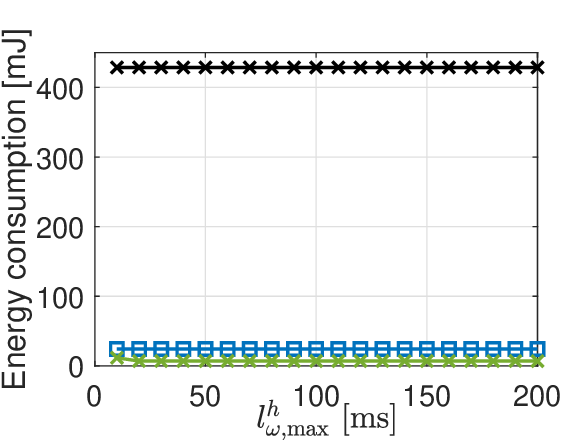}
         }
         \hspace{-1mm} 
\subfigure[$\ell^h_{\omega,\max}=$50\,ms \label{f:tot_energy_motorwaya}]{   \includegraphics[width=0.31\columnwidth]{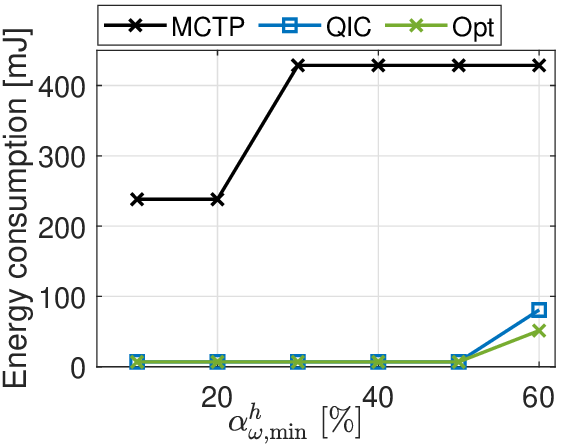}  
    }\hspace{-1mm}
\subfigure[$\alpha^h_{\omega,\min}=$50$ \%$\label{f:tot_energy_motorwayb}]{
\includegraphics[width=0.31\columnwidth]{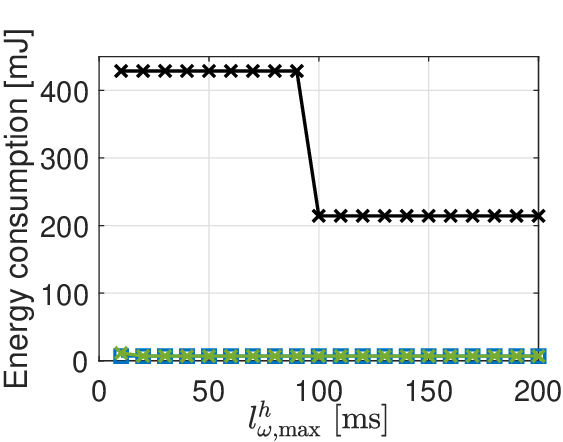}
         }
          \vspace*{-4mm}
    \caption{Small-scale scenario: Total energy consumption obtained through MCTP, QIC, and Opt, in the \texttt{Sunny}  (a,b), \texttt{Night} (c,d), and \texttt{Motorway} (e,f) context, as the target inference  latency and accuracy quantiles vary.}
    \centering
        \subfigure[Average energy\label{f:multi_energy}]{
         \includegraphics[width=0.45\columnwidth]{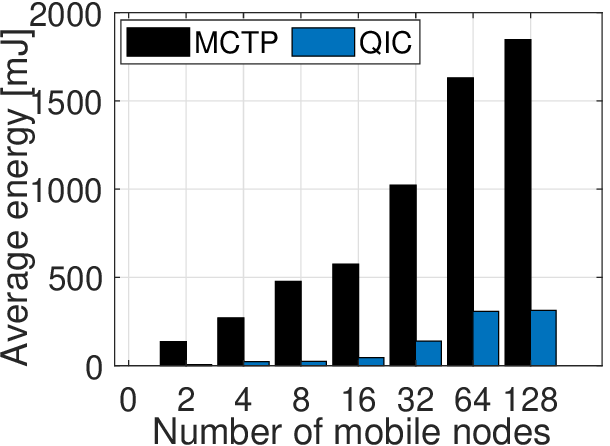}
         }\hspace{-0mm}
          \subfigure[Churn rate\label{f:multi_churn}]{
         \includegraphics[width=0.45\columnwidth]{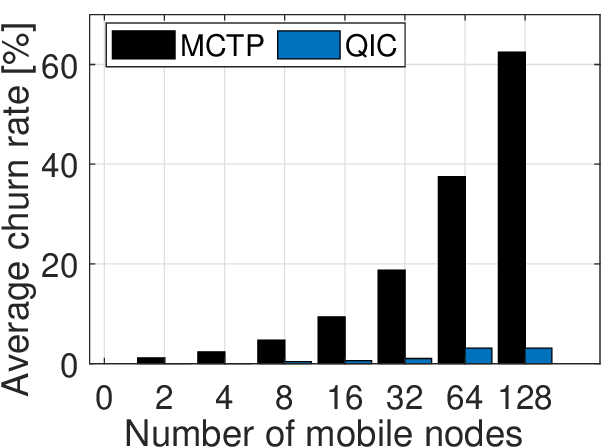}
         }
             \subfigure[Average latency\label{f:multi_latency}]{
      \includegraphics[width=0.45\columnwidth]{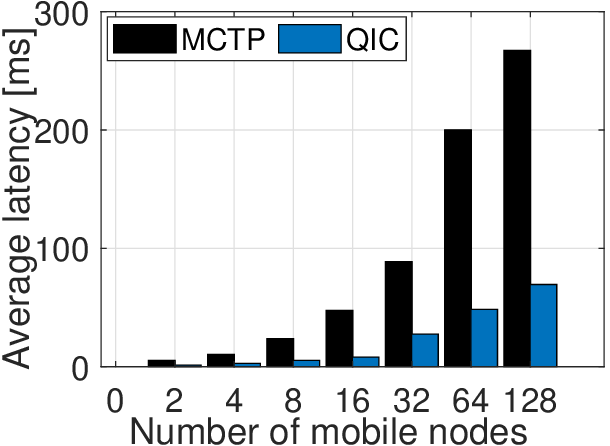}  
    }\hspace{-0mm}
      \subfigure[Average accuracy\label{f:multi_accuracy}]{
      \includegraphics[width=0.45\columnwidth]{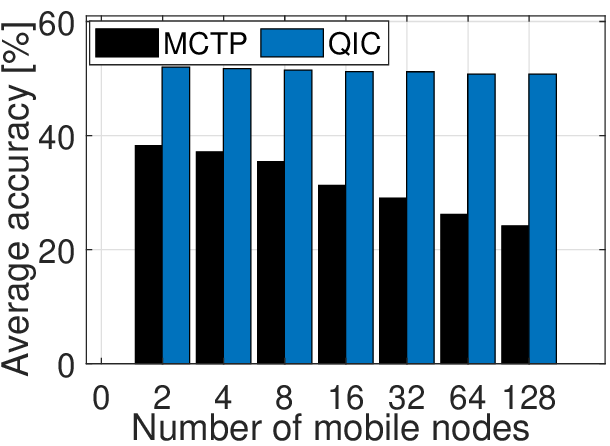}  
    }
\vspace{-4mm}
    \caption{Large-scale scenario:  Performance as the number of mobile nodes increases   ($\ell^h_{\omega,\max}${=}50\,ms,  $\alpha^h_{\omega,\min}${=}50 \%,  $\omega=0.9$).}
    \label{f:multi}
    \vspace{-4mm}
\end{figure*}

Parameters are set as listed in Table~\ref{t:sim}.    
As mentioned, in both the small-scale and large-scale scenario,  the  available radio resources and the MCS index that can be used by each  mobile node are changing over time.

{\bf Small-scale scenario.} 
The scenario, depicted in Fig.\,\ref{f:sample_scen}, includes  4 data sources, 3 mobile nodes, 2 ESs, and 3 applications. We associate each mobile node with one application and a specific context, and  investigate the impact on  energy consumption of accuracy and latency constraints. 
We begin by looking at the performance in the case of the mobile node with the \texttt{Sunny} context. In Fig.~\ref{f:tot_energy_sunnya}, we fix the latency target to~$\ell^h_{\omega,\max}{=}50\text{ ms}$ with quantile $\omega{=}0.9$ %
and vary the accuracy target~$\alpha^h_{\omega,\min}$; as one might expect, a tighter accuracy target results in a larger energy consumption for all strategies. More importantly, QIC greatly outperforms MCTP, yielding savings that exceed 25\%, and it almost always matches the optimum. In Fig.~\ref{f:tot_energy_sunnyb}, we fix the accuracy target to~$\alpha^h_{\omega,\min}{=}50\%$ with quantile $\omega{=}0.9$ and change the latency target  (again, MCTP  cannot meet the target accuracy and latency  with higher $\omega$'s). Besides noticing that shorter latency results in higher energy consumption,  remarkably, QIC can achieve the same performance as the optimum, except when latency constraints are very tight, and consistently outperforms MCTP, on average, by 80\%.  
The same behavior can be observed for the \texttt{Night} and \texttt{Motorway} contexts (Figs.~\ref{f:tot_energy_nighta}--\ref{f:tot_energy_nightb} and Figs.~\ref{f:tot_energy_motorwaya}--\ref{f:tot_energy_motorwayb}, resp.).   Since the maximum accuracy for  \texttt{Night} and \texttt{Motorway} is now limited to the less stringent requirement of $60\%$, the difference between the different schemes is occasionally slightly smaller than in the \texttt{Sunny} context.  Nevertheless, QIC consistently makes optimal or near-optimal decisions, while MCTP always incurs  25\% higher energy consumption relatively  to QIC.

{\bf Large-scale scenario.} Next, we  apply QIC to a  scenario with multiple mobile nodes  and 10 ESs. 
Context  (\texttt{Sunny}, \texttt{Night}, and \texttt{Motorway}) and applications are  uniformly assigned to  mobile nodes. 
We set $\ell^h_{\omega,\max}{=}50$\,ms, $\alpha^h_{\omega,\min}{=}50\%$, and $\omega{=}0.9$ for all applications (for larger $\omega$'s MCTP cannot meet the requirements).  
Fig.~\ref{f:multi} compares  QIC  to  the benchmark, in terms of average performance of  energy  and churn rate (ratio of the no.\,of mobile nodes that fail to meet the application requirements to the total number of mobile nodes in the system), with the average value computed over the entire duration of the radio link measurement trace and the number of mobile nodes. Figs.\,\ref{f:multi_energy}--\ref{f:multi_churn} underline the increase of the above metrics as the number of mobile nodes, hence the load on the edge servers, grows. 
More interestingly,  QIC has excellent energy performance and always meets the application target accuracy and latency quantiles, even with high number of mobile nodes, whereas MCTP fails to meet these requirements for more than 8 mobile nodes. 
Overall, QIC results in more than 80\% reduction in  energy consumption and on average 70\% higher number of mobile nodes meeting their application requirements as compared to MCTP.  
Figs.~\ref{f:multi_latency}--\ref{f:multi_accuracy} shed light on  MCTP's high churn rate, by depicting the average latency and accuracy for both QIC and MCTP. It is clear that for more than 8 mobile nodes, MCTP violates the latency requirement and it does so by a large margin.

\section{Conclusions}

We targeted dynamic scenarios where multiple DNN-based applications can leverage multiple  data sources for their inference task. Given  DNN's architecture with multiple  stems and branches that can be dynamically selected, 
we  proposed QIC to jointly choose (i) the input data sources and (ii) the DNN sections to use, and  (iii) the  nodes where each stem and branch is deployed, along with (iv) the resources to use therein. We proved QIC's polynomial worst-case time complexity and,  using  a dynamic DNN architecture, a real-world dataset and radio link measurements, showed that QIC closely matches the optimum and outperforms its benchmarks by over 80\%.

\section*{Acknowledgments}
This work was supported by the European Union through Grant No.\,101139266 (6G-INTENSE project) and the Next Generation EU under the Italian NRRP, M4C2, Investment 1.3, CUP E13C22001870001, PE00000001 - program “RESTART”. 
\vspace{-2mm}

\bibliographystyle{IEEEtran}
\bibliography{refs_central}
\end{document}